# Continuous online sequence learning with an unsupervised neural network model

Yuwei Cui*, Subutai Ahmad, and Jeff Hawkins
Numenta, Inc,
Redwood City, California, USA

*Corresponding author
Emails: ycui@numenta.com, sahmad@numenta.com, jhawkins@numenta.com

# Continuous online sequence learning with an unsupervised neural network model

Yuwei Cui, Subutai Ahmad, and Jeff Hawkins
Numenta, Inc, Redwood City, California, United States of America

## Abstract

The ability to recognize and predict temporal sequences of sensory inputs is vital for survival in natural environments. Based on many known properties of cortical neurons, hierarchical temporal memory (HTM) sequence memory is recently proposed as a theoretical framework for sequence learning in the cortex. In this paper, we analyze properties of HTM sequence memory and apply it to sequence learning and prediction problems with streaming data. We show the model is able to continuously learn a large number of variable-order temporal sequences using an unsupervised Hebbian-like learning rule. The sparse temporal codes formed by the model can robustly handle branching temporal sequences by maintaining multiple predictions until there is sufficient disambiguating evidence. We compare the HTM sequence memory with other sequence learning algorithms, including statistical methods: autoregressive integrated moving average (ARIMA), feedforward neural networks: online sequential extreme learning machine (ELM), and recurrent neural networks: long short-term memory (LSTM) and echo-state networks (ESN), on sequence prediction problems with both artificial and real-world data. The HTM model achieves comparable accuracy to other state-of-the-art algorithms. The model also exhibits properties that are critical for sequence learning, including continuous online learning, the ability to handle multiple predictions and branching sequences with high order statistics, robustness to sensor noise and fault tolerance, and good performance without task-specific hyper-parameters tuning. Therefore the HTM sequence memory not only advances our understanding of how the brain may solve the sequence learning problem, but is also applicable to a wide range of real-world problems such as discrete and continuous sequence prediction, anomaly detection, and sequence classification.

## 1. Introduction

In natural environments, the cortex continuously processes streams of sensory information and builds a rich spatio-temporal model of the world. The ability to recognize and predict ordered temporal sequences is critical to almost every function of the brain, including speech recognition, active tactile perception, and natural vision. Neuroimaging studies have demonstrated that multiple cortical regions are involved in temporal sequence processing (Clegg et al., 1998; Mauk and Buonomano, 2004). Recent neurophysiology studies have shown that even neurons in primary visual cortex can learn to recognize and predict spatiotemporal sequences (Xu et al., 2012; Gavornik and Bear, 2014) and that neurons in primary visual and auditory cortex exhibit sequence sensitivity (Brosch and Schreiner, 2000; Nikolić et al., 2009). These studies suggest that sequence learning is an important problem that is solved by many cortical regions.

Machine learning researchers have also extensively studied sequence learning independently of neuroscience. Statistical models, such as hidden-markov models (HMM)(Rabiner and Juang, 1986; Fine et al., 1998) and autoregressive integrated moving average (ARIMA)(Durbin and Koopman, 2012) have been developed for temporal pattern recognition and time-series prediction respectively. A variety of neural network models have been proposed to model sequential data. Feedforward networks, such as time delay neural networks (TDNN), have been used to model sequential data by adding a set of delays to the input (Waibel et al., 1989). Recurrent neural networks can model sequence structure with recurrent lateral connections and process the data sequentially one record at a time. For example, long short-term memory (LSTM) has the ability to selectively pass information across time and can model very long term dependencies using gating mechanisms (Hochreiter and Schmidhuber, 1997) and gives impressive performance on a wide variety of real-world problems (Lipton et al., 2015). Echo state network (ESN) uses a randomly connected recurrent network as a dynamics reservoir and model a sequence as trainable linear combination of these response signals (Jaeger and Haas, 2004).

Can machine learning algorithms gain any insight from cortical algorithms? The current state-of-the-art statistical and machine-learning algorithms achieve impressive prediction accuracy on benchmark problems. However, most time-series prediction benchmarks do not focus on model performance in dynamic, non-stationary scenarios. Benchmarks typically have separate training and testing datasets, where the underlying assumption is that the test data share similar statistics as the training data (Crone et al., 2011; Ben Taieb et al., 2012). In contrast, sequence learning in the brain has to occur continuously to deal with the noisy, constantly changing streams of sensory inputs. Notably, with the increasing availability of streaming data, there is also an increasing demand for online sequence algorithms that can handle complex, noisy data streams. Therefore, reverse-engineering the computational principles used in the brain could offer additional insights into the sequence learning problem that lies at the heart of many machine learning applications.

The exact neural mechanisms underlying sequence memory in the brain remain unknown but biologically plausible models based on spiking neurons have been studied. For example, (Rao and Sejnowski, 2001) showed that spike-time-dependent plasticity rules can lead to predictive sequence learning in recurrent neocortical circuits. Spiking recurrent network models have been shown to recognize and recall precisely timed sequences of inputs using supervised learning rules (Ponulak and Kasiński, 2010; Brea et al., 2013). These studies demonstrate that certain limited types of sequence learning can be solved with biologically plausible mechanisms. However, only a few practical sequence learning applications use spiking network models as these models only recognize relatively simple and limited types of sequences. These models also do not match performance of non-biological statistical and machine learning approaches on real-world problems.

In this paper we present a comparative study of HTM sequence memory, a detailed model of sequence learning in the cortex (Hawkins and Ahmad, 2016). The HTM neuron model incorporates many recently discovered properties of pyramidal cells and active dendrites (Antic et al., 2010; Major et al., 2013). Complex sequences are represented using sparse distributed temporal codes (Kanerva, 1988; Ahmad and Hawkins, 2016) and the network is trained using an online unsupervised Hebbian-style learning rule. The algorithms have been applied to many practical problems, including discrete and continuous sequence prediction, anomaly detection (Lavin and Ahmad, 2015), and sequence recognition and classification.

We compare HTM sequence memory with four popular statistical and machine learning techniques, including ARIMA, a statistical method for time-series forecasting (Durbin and Koopman, 2012); extreme learning machine (ELM), a feedforward network with sequential online learning (Huang et al., 2006; Liang et al., 2006) and two recurrent networks LSTM and ESN. We show that HTM sequence memory achieves comparable prediction accuracy to these other techniques. In addition it exhibits a set of features that is desirable for real-world sequence learning from streaming data. We demonstrate that HTM networks learns complex high-order sequences from data streams, rapidly adapts to changing statistics in the data, naturally handles multiple predictions and branching sequences and exhibits high tolerance to system faults.

The paper is organized as follows. In section 2, we discuss a list of desired properties of sequence learning algorithms for real-time streaming data analysis. In section 3, we introduce the HTM temporal memory model. In sections 4 and 5, we apply the HTM temporal memory and other sequence learning algorithms to discrete artificial data and continuous real world data respectively. Discussion and conclusions are given in section 6.

## 2. Criteria for a good sequence learning algorithm

With the increasing availability of streaming data, there is an increasing demand for online sequence learning algorithms. Here, a data stream is an ordered sequence of data records that must be processed in real-time using limited computing and storage capabilities. In the field of data stream mining, the goal is to extract knowledge from continuous data streams such as computer network traffic, sensor data, financial transactions, etc. (Domingos and Hulten, 2000; Gaber et al., 2005; Gama, 2010), which often have changing statistics (non-stationary) (Sayed-Mouchaweh and Lughofer, 2012). Real-world sequence learning from such complex, noisy data streams requires many other properties in addition to prediction accuracy. This stands in contrast to many machine learning algorithms, which are developed to optimize performance on static datasets, and lack the flexibility to solve real-time streaming data analysis tasks.

In contrast to these algorithms, the cortex solves the sequence learning problem in a drastically different way. Rather than achieving optimal performance for a specific problem (e.g., through gradient-based optimization), the cortex learns continuously from noisy sensory input streams and quickly adapts to the changing statistics of the data. When information is insufficient or ambiguous, the cortex can make multiple plausible predictions given the available sensory information.

Real-time sequence learning from data streams presents unique challenges for machine learning algorithms. In addition to prediction accuracy, below we list a set of criteria that applies to both biological systems and real-world streaming applications.

*1) Continuous learning*

Continuous data streams often have changing statistics. As a result, the algorithm needs to continuously learn from the data streams and rapidly adapt to changes. This property is important for processing continuous real-time sensory streams, but has not been well studied in machine learning. For real-time data stream analysis, it is much valuable if the algorithm can recognize and learn new patterns rapidly.

Machine learning algorithms can be classified into batch or online learning algorithms. Both types of algorithms can be adopted for continuous learning applications. To apply a batch-learning algorithm to continuous data stream analysis, one needs to keep a buffered dataset of past data records. The model is retrained at regular intervals as the statistics of the data can change over time. The batch-training paradigm potentially requires significant computing and storage resources, particularly in situations where the data velocity is high. In contrast, online sequential algorithms can learn sequences in a single-pass and do not require a buffered dataset.

*2) High-order predictions*

Real-world sequences contain contextual dependencies that span multiple time steps, i.e. the ability to make high-order predictions. The term "order" refers to Markov order, specifically the minimum number of previous time steps the algorithm needs to consider in order to make accurate predictions. An ideal algorithm should learn the order automatically and efficiently.

*3) Multiple simultaneous predictions*

For a given temporal context, there could be multiple possible future outcomes. With real-world data, it is often insufficient to only consider the single best prediction when information is ambiguous. A good sequence learning algorithm should be able to make multiple predictions simultaneously and evaluate the likelihood of each prediction online. This requires the algorithm to output a distribution of possible future outcomes. This property is present in HMMs (Rabiner and Juang, 1986) and generative recurrent neural network models (Hochreiter and Schmidhuber, 1997), but not in other approaches like ARIMA, which are limited to maximum likelihood prediction.

*4) Noise robustness and fault tolerance*

Real world sequence learning deals with noisy data sources where sensor noise, data transmission errors and inherent device limitations frequently result in inaccurate or missing data. A good sequence learning algorithm should exhibit robustness to noise in the inputs.

The algorithm should also be able to learn properly in the event of system faults, such as loss of synapses and neurons in a neural network. The property of fault tolerance and robustness to failure is present in the brain. This property is important for the development of next-generation neuromorphic processors (Tran et al., 2011). Noise robustness and fault tolerance ensures flexibility and wide applicability of the algorithm to a wide variety of problems.

*5) No hyperparameter tuning*

Learning in the cortex is extremely robust for a wide range of problems. In contrast, most machine-learning algorithms require optimizing a set of hyperparameters for each task. It typically involves searching through a manually specified subset of the hyperparameter space, guided by performance metrics on a cross-validation dataset. Hyperparameter tuning presents a major challenge for applications that require a high degree of automation, like data stream mining. An ideal algorithm should have acceptable performance on a wide range of problems without any task-specific hyperparameter tuning.

Many of the existing machine learning techniques demonstrate these properties to various degrees. A truly flexible and powerful system for streaming analytics would meet all of them. In the rest of the paper, we will compare HTM sequence memory with other common sequence learning algorithms (ARIMA, ELM, ESN and LSTM) on various tasks using the above criteria.

## 3. HTM sequence memory

In this section we describe the computational details of HTM sequence memory. We first describe our neuron model. We then describe the representation of high order sequences, followed by a formal description of our learning rules. We point out some of the relevant neuroscience experimental evidence in our description, but a detailed mapping to the biology can be found in (Hawkins and Ahmad, 2016).

### 3.1. HTM neuron model

The HTM neuron (Fig. 1B) implements non-linear synaptic integration inspired by recent neuroscience findings regarding the function of cortical neurons and dendrites (Spruston, 2008; Major et al., 2013). Each neuron in the network contains two separate zones: a proximal zone containing a single dendritic segment and a distal zone containing a set of independent dendritic segments. Each segment maintains a set of synapses. The source of the synapses is different depending on the zone (Fig. 1B). Proximal synapses represent feed-forward inputs into the layer whereas distal synapses represent lateral connections within a layer and feedback connections from a higher region. In this paper, we only consider a single region and ignore feedback connections.

Each distal dendritic segment contains a set of lateral synaptic connections from other neurons within the layer. A segment becomes active if the number of simultaneously active connections exceeds a threshold. An active segment does not cause the cell to fire but instead causes the cell to enter a depolarized state, which we call the "predictive state". In this way each segment detects a particular temporal context and makes predictions based on that context. Each neuron can be in one of three internal states: an active state, a predictive state, or a non-active state. The output of the neuron is always binary: it is either active or not.

The above neuron model is inspired by a large number of recent experimental findings that suggest neurons do not perform a simple weighted sum of their inputs and fire based on that sum (Polsky et al., 2004; Smith et al., 2013), as in most neural network models (McFarland et al., 2013; Schmidhuber, 2014; LeCun et al., 2015). Instead, dendritic branches are active processing elements. The activation of several synapses within close spatial and temporal proximity on a dendritic branch can initiate a local NMDA spike, which then cause a significant and sustained depolarization of the cell body (Antic et al., 2010; Major et al., 2013).

### 3.2. Two separate sparse representations

The HTM network consists of a layer of HTM neurons organized into a set of columns (Fig. 1A). The network represents high-order sequences using a composition of two separate sparse representations. At any time, both the current feedforward input and the previous sequence context are simultaneously represented using sparse distributed representations.

The first representation is at the column level. We assume that all neurons within a column detect identical feed-forward input patterns on their proximal dendrites (Mountcastle, 1997; Buxhoeveden, 2002). Through an inter-columnar inhibition mechanism, each input element is encoded as a sparse distributed activation of columns at any point in time. At any time, the top 2% columns that receive most active feedforward inputs are activated.

The second representation is at the level of individual cells within these columns. At any given point a subset of cells in the active columns will represent information regarding the learned temporal context of the current pattern. These cells in turn lead to predictions of the upcoming input through lateral projections to other cells within the same network. The predictive state of a cell controls inhibition within a column. If a column contains predicted cells and later receives sufficient feed-forward input, these cells become active and inhibit others within that column. If there were no cells in the predicted state, all cells within the column become active.

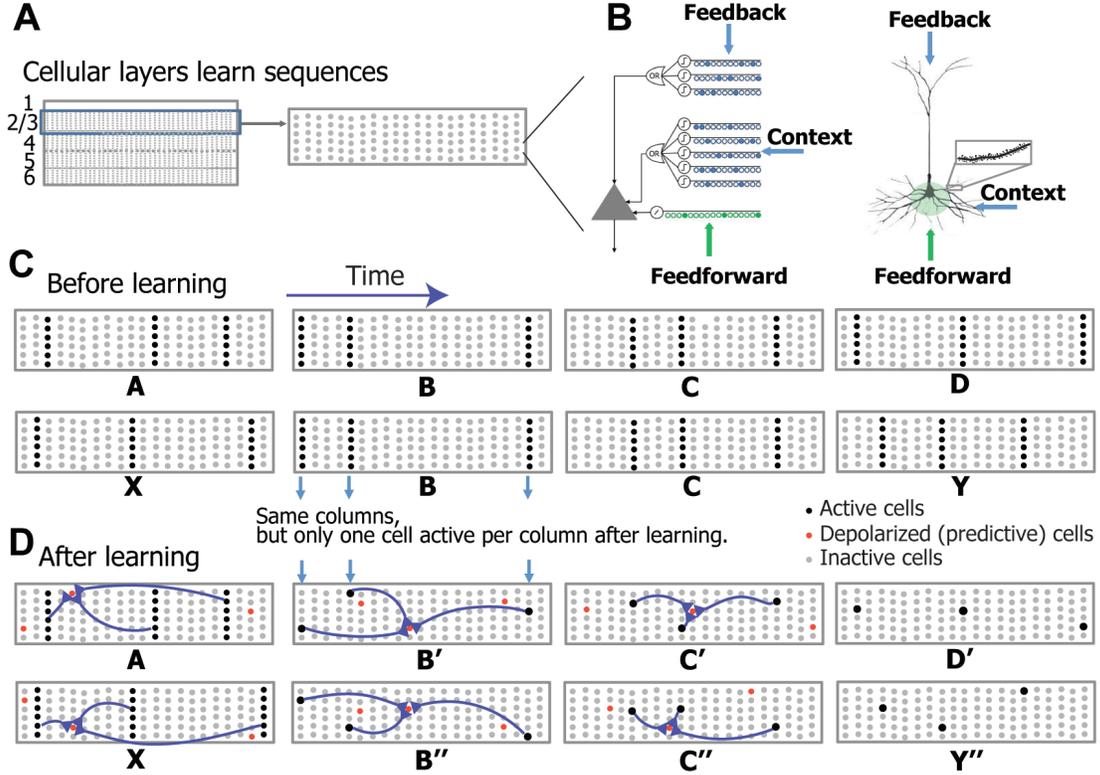

**Figure 1** The HTM sequence memory model. **A**. The cortex is organized into 6 cellular layers. Each cellular layer consists of a set of mini-columns, with each mini-column containing multiple cells. **B**. An HTM neuron (left) has three distinct dendritic integration zones, corresponding to different parts of the dendritic tree of pyramidal neurons (right). An HTM neuron models dendrites and NMDA spikes asç an array of coincident detectors, each with a set of synapses. The co-activation of a set of synapses on a distal dendrite will cause an NMDA spike and depolarize the soma (predicted state). **C-D**. Learning high-order Markov sequences with shared subsequences (ABCD vs. XBCY). Each sequence element invokes a sparse set of mini-columns due to inter-column inhibition. **C**. Prior to learning the sequences, all the cells in a mini-column become active. **D**. After learning, cells that are depolarized through lateral connections become active faster and prevent other cells in the same column from firing through intra-column inhibition. The model maintains two simultaneous representations: one at the mini-column level representing the current feedforward input, and the other at individual cell level representing the context of the input. Because different cells respond to "C" in the two sequences (C' and C''), they can invoke the correct high-order prediction of either D or Y.

To illustrate the intuition behind these representations, consider two abstract sequences **A-B-C-D** and **X-B-C-Y** (Fig. 1C-D). In this example remembering that the sequence started with **A** or **X** is required to make the correct prediction following "**C**". The current inputs are represented by the subset of columns that contains active cells (*black*, Fig. 1C-D). This set of active columns does not depend on temporal context, just on the current input. After learning, different cells in this subset of columns will be active depending on predications based on the past context (**B'** vs. **B''**, **C'** vs. **C''**, Fig. 1D). These cells then lead to predictions of the element following **C** (**D** or **Y**) based on the set of cells containing lateral connections to columns representing **C**.

This dual representation paradigm leads to a number of interesting properties. First, the use of sparse representations allows the model to make multiple predictions simultaneously. For example, if we present input "**B**" to the network without any context, all cells in columns representing the "**B**" input will fire, which leads to a prediction of both **C'** and **C''**. Second, because information is stored by co-activation of multiple cells in a distributed manner, the model is naturally robust to both noise in the input and system faults such as loss of neurons and synapses. A detailed discussion on this topic can be found in (Hawkins and Ahmad, 2016).

### 3.3. HTM activation and learning rules

The previous sections provided an intuitive description of network behavior. In this section we describe the formal activation and learning rules for the HTM network. Consider a network with $N$ columns and $M$ neurons per column; we denote the activation state at time step $t$ with an $M \times N$ binary matrix $\mathbf{A}^t$, where $a_{ij}^t$ is the activation state of the $i$'th cell in the $j$'th column. Similarly, an $M \times N$ binary matrix $\mathbf{\Pi}^t$ denotes cells in a predictive state at time $t$, where $\pi_{ij}^t$ is the predictive state of the $i$'th cell in the $j$'th column. We model each synapse with a scalar permanence value, and consider a synapse connected if its permanence value is above a connection threshold. We use an $M \times N$ matrix $\mathbf{D}_{ij}^{d}$ to denote the permanence of $d$'th segment of the $i$'th cell in the $j$'th column. The synaptic permanence matrix is bounded between 0 and 1. We use a binary matrix $\widetilde{\mathbf{D}}_{ij}^{d}$ to denote only the connected synapses. The network can be initialized such that each segment contains a set of potential synapses (i.e. with non-zero permanence value) to a randomly chosen subset of cells in the layer. To speed up simulation, instead of explicitly initializing a complete set of synapses across every segment and every cell, we greedily create segments at run time (see Appendix).

The predictive state of the neuron is handled as follows: if a dendritic segment receives enough input, it becomes active and subsequently depolarizes the cell body without causing an immediate spike. Mathematically, the predictive state at time step $t$ is calculated as follows:

$$\pi_{ij}^t = \begin{cases} 1 \text{ if } \exists_d \left\| \widetilde{\mathbf{D}}_{ij}^d \circ \mathbf{A}^t \right\|_1 > \theta \\ 0 \text{ otherwise} \end{cases} \quad (1)$$

Threshold $\theta$ represents the segment activation threshold and $\circ$ represents element-wise multiplication. Since the distal synapses receive inputs from previously active cells in the same layer, it contains contextual information about future inputs (Fig. 1B).

At any time, an inter-columnar inhibitory process select a sparse set of columns that best match the current feed forward input pattern. We calculate the number of active proximal synapses for each column, and activate the top 2% of the columns that receive the most synaptic inputs. We denote this set as $\mathbf{W}^t$. The proximal synapses were initialized such that each column is randomly connected to 50% of the inputs. Since we focused sequence learning in this paper, the proximal synapses were fixed during learning. In principle, the proximal synapses can also adapt continuously during learning according to a spatial competitive learning rule (Hawkins et al., 2011; Mnatzaganian et al., 2016).

Neurons in the predictive state (i.e. depolarized) will have competitive advantage over other neurons receiving the same feed-forward inputs. Specifically, a depolarized cell fires faster than other non-depolarized cells if it subsequently receives sufficient feed-forward input. By firing faster, it prevents neighboring cells in the same column from activating with intra-column inhibition. The active state for each cell is calculated as follows:

$$a_{ij}^t = \begin{cases} 1 \text{ if } j \in \mathbf{W}^t \text{ and } \pi_{ij}^{t-1} = 1 \\ 1 \text{ if } j \in \mathbf{W}^t \text{ and } \sum_i \pi_{ij}^{t-1} = 0 \\ 0 \text{ otherwise} \end{cases} \quad (2)$$

The first conditional expression of Eq. 2 represents a cell in a winning column becoming active if it was in a predictive state during the preceding time step. If none of the cells in a winning column are in a predictive state, all cells in that column become active, as in the second conditional of Eq. 2.

The lateral connections in the sequence memory model are learned using a Hebbian-like rule. Specifically, if a cell is depolarized and subsequently becomes active, we reinforce the dendritic segment that caused the depolarization. If no cell in an active column is predicted, we select the cell with the most activated segment and reinforce that segment. Reinforcement of a dendritic segment involves decreasing permanence values of inactive synapses by a small value $p^-$ and increasing the permanence for active synapses by a larger value $p^+$:

$$\Delta \mathbf{D}_{ij}^d = p^+ \dot{\mathbf{D}}_{ij}^d \circ \mathbf{A}^{t-1} - p^- \dot{\mathbf{D}}_{ij}^d \circ (\mathbf{1} - \mathbf{A}^{t-1}) \quad (3)$$

$\dot{\mathbf{D}}_{ij}^d$ denotes a binary matrix containing only the positive entries in $\mathbf{D}_{ij}^d$. We also apply a very small decay to active segments of cells that did not become active, mimicking the effect of long-term depression (Massey and Bashir, 2007):

$$\Delta \mathbf{D}_{ij}^d = p^{--} \dot{\mathbf{D}}_{ij}^d \text{ where } a_{ij}^t \\ = 0 \text{ and } \left\| \widetilde{\mathbf{D}}_{ij}^d \circ \mathbf{A}^{t-1} \right\|_1 > \theta \quad (4)$$

where $p^{--} \ll p^-$.

The learning rule is inspired by neuroscience studies of activity-dependent synaptogenesis (Zito and Svoboda, 2002), which showed that the adult cortex generates new synapses in response to sensory activity rapidly. The mathematical formula we chose captured this Hebbian synaptogenesis learning rule. We did not derive the rule by implementing gradient descent on a cost function. There could be other mathematical formulations that give similar or better results.

A complete set of parameters and further implementation details can be found in the appendix. These parameters were set based on properties of sparse distributed representations (Ahmad and Hawkins, 2016). Notably, we used the same set of parameters for all the different types of sequence learning tasks in this paper.

### 3.4. SDR encoder and classifier

The HTM sequence memory operates with sparse distributed representations (SDRs) internally. To apply HTM to real-world sequence learning problems, we need to first convert the original data to SDRs using an encoder. We have created a variety of encoders to deal with different data types (Purdy, 2016). In this paper, we used a random SDR encoder for categorical data, and used scalar and date-time encoders for the taxi passenger prediction experiment.

To decode prediction values from the output SDRs of HTM, we considered two classifiers: a simple classifier based on SDR overlaps and a maximum-likelihood classifier. For the single-step discrete sequence prediction task, we computed the overlap of the predicted cells with the SDRs of all observed elements and selected the one with the highest overlap. For the continuous scalar value prediction task, we divided the whole range of scalar value into 22 disjoint buckets, and used a single layer feedforward classification network. Given a large array of cell activation pattern $\mathbf{x}$, the classification network computes a probability distribution over all possible classes using a softmax activation function (Bridle, 1989). There are as many output units as the number of possible classes. The jth output unit receives a weighted summation of all the inputs,

$$a_j = \sum_{i=1}^{N} w_{ij} x_i \qquad (5)$$

$w_{ij}$ is the connection weight from the ith input neuron to the jth output neuron. The estimated class probability is given by the activation level of the output units,

$$y_k = P(C_k|\mathbf{x}) = \frac{e^{a_k}}{\sum_{i=1}^{K} e^{a_k}} \qquad (6)$$

Using a maximum likelihood optimization, we derived the learning rule for the weight matrix **w**.

$$\Delta w_{ij} = \lambda(y_j - z_j) x_i \qquad (7)$$

$z_j$ is the observed (target) distribution and $\lambda$ is the learning rate. Note that since **x** is highly sparse, we only need to update a very small fraction of the weight matrix at any time. Therefore the learning algorithm for the classifier is fast despite the high dimensionality of the weight matrix.

## 4. High-order sequence prediction with artificial data

We conducted experiments to test whether the HTM sequence memory model, online sequential extreme learning machine (OS-ELM) and the LSTM network are able to learn high-order sequences in an online manner, recover after modification to the sequences, and make multiple predictions simultaneously. LSTM represents the state-of-the-art recurrent neural network model for sequence learning tasks (Hochreiter and Schmidhuber, 1997; Graves, 2012). OS-ELM is a feedforward neural network model that is widely used for time-series predictions (Huang et al., 2011; Wang and Han, 2014).

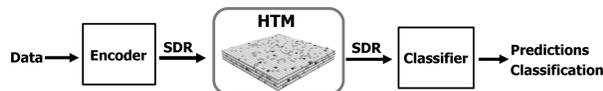

**Figure 2** Functional steps for using HTM on real-world sequence learning tasks.

### 4.1. Continuous online learning from streaming data

We created a discrete high-order temporal sequence dataset. Sequences are designed such that any learning algorithm would have to maintain context of at least the first two elements of each sequence in order to correctly predict the last element of the sequence (Figure 3). We used the sequence dataset in a continuous streaming scenario (Fig. 3C). At the beginning of a trial, we randomly chose a sequence from the dataset and sequentially presented each of its elements. At the end of each sequence, we presented a single noise element to the model. The noise element is randomly chosen from a large set of 50,000 noise symbols (not used among the set of sequences). This is a difficult learning problem, since sequences are embedded in random noise; the start and end points are not marked. The set of noise symbols is large so the algorithm cannot learn every possible noise transition. We tested the algorithms for predicting the last element of each sequence continuously as the algorithm observed a stream of sequences, and reported the percentage of correct predictions over time.

We encoded each symbol in the sequence as a random SDR for HTM sequence memory, with 40 randomly chosen active bits in a vector of 2048 bits. This SDR representation matches the internal representation used in HTM, which has 2048 columns with 40 active at any time (see Implementation Details). We initially tried to use the same SDR encoding for ELM and LSTM. This high dimensional representation does not work well for ELM and LSTM due to the large number of parameters required. Instead, we used a random real-valued dense distributed representation for ELM and LSTM. Each symbol is encoded as a 25-dimensional vector with each dimension's value randomly chosen from [-1, 1][1]. We chose this encoding format because it both gives better accuracy and has large representational capacity, which is required for streaming data analysis. Similar dense distributed representations are commonly used for LSTM in natural language processing applications (Mikolov et al., 2013).

Since the sequences are presented in a streaming fashion and predictions are required continuously, this task represents a continuous online learning problem. The HTM sequence memory is naturally suitable for online learning as it learns from every new data point and does not require the data stream to be broken up into predefined chunks. ELM also has a well-established online sequential learning model (Liang et al., 2006). Online sequential algorithm, such as real-time recurrent learning (RTRL) has been proposed for LSTM in the past, (Williams and Zipser, 1989; Hochreiter and Schmidhuber, 1997). However, most LSTM applications used batch learning due to the high computational cost of RTRL (Jaeger, 2002). We use two variants of LSTM networks for this task. First, we retrained an LSTM network at regular intervals on a buffered dataset of the previous time steps using a variant of the resilient backpropagation algorithm until convergence (Igel and Hüsken, 2003). The experiments include several LSTM models with varying buffer sizes. Second, we trained a LSTM network with online truncated back-propagation through time (BPTT) (Williams and Peng, 1990). At each time point, we calculated the gradient using BPTT over the last 100 elements and adjusted the parameters along the gradient by a small amount.

We tested sequences with either single or multiple possible endings (Fig. 3A-B). To quantify model performance, we classified the state of the model before presenting the last element of each sequence to retrieve the top $K$ predictions, where $K = 1$ for the single prediction case and $K = 2$ or 4 for the multiple predictions case. We considered the prediction correct if the actual last element was among the top $K$ predictions of the model. Since these are online learning tasks, there are no separate training and test phases. Instead, we continuously report the prediction accuracy of the end of each sequence before the model has seen it.

---

[1] We manually tuned the number of dimensions and found that 25 dimensions gave the best performance on our tasks.

In the single prediction experiment (Fig. 4, left of the black dashed line), each sequence in the dataset has only one possible ending given its high-order context (Fig. 3A). The HTM sequence memory quickly achieves perfect prediction accuracy on this task (Fig. 4, *red*). Given a large enough learning window, LSTM also learns to predict the high-order sequences (Fig. 4, *green*). Despite comparable model performance, TM and LSTM use the data in different ways: LSTM requires many passes over the learning window each time it is retrained to perform gradient-descent optimization, whereas HTM only needs to see each element once (one-pass learning). LSTM also takes longer than HTM to achieve perfect accuracy; we speculate that since LSTM optimizes over all transitions in the data stream, including the random ones between sequences, it is initially overfitting on the training data. Online LSTM and ELM are also trained in an online, sequential fashion similar to HTM. But both algorithms require keeping a short history buffer of the past elements. ELM learned the sequences slower than HTM, and never achieved perfect performance (Fig. 4, *blue*). Online LSTM has the best performance initially, but does not achieve perfect performance in the end. HTM and LSTM are the only algorithms to achieve perfect prediction accuracy on this task.

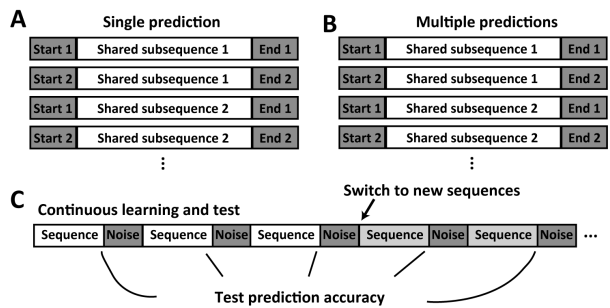

**Figure 3** Design of the high-order sequence prediction task. **A**. Structure of high order sequences with shared subsequences. **B**. High order sequences with multiple possible endings. **C**. Stream of sequences with noise between sequences. Both learning and testing occur continuously. After the model learned one set of sequences, we switched to a new set of sequences with contradictory endings to test the adaptation to changes in the data stream.

### 4.2. Adaptation to changes in the data stream

Once the models have achieved stable performance, we altered the dataset by swapping the last elements of pairs of high-order sequences (Fig. 4, black dashed line). This forces the model to forget the old sequences and subsequently learn the new ones. HTM sequence memory and online LSTM quickly recover from the modification. In contrast, it takes a long time for batch LSTM and ELM to recover from the modification as its buffered dataset contains contradictory information before and after the modification. Although using a smaller learning window can speed up the recovery (Fig. 4, *blue; purple*), it also causes worse prediction performance due to limited number of training samples.

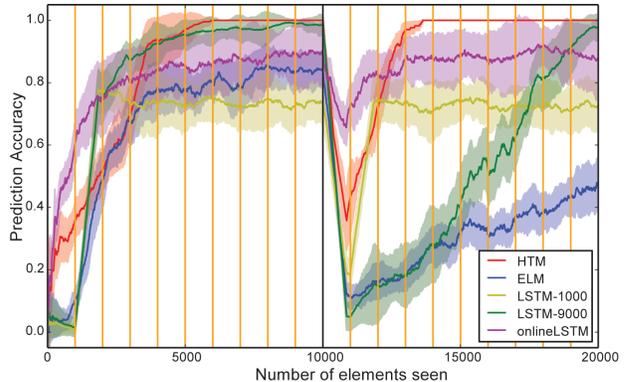

**Figure 4** Prediction accuracy of HTM (*red*) and LSTM (*blue, green, purple*) on an artificial dataset. The dataset contains four $6^{th}$ order sequences and four $7^{th}$ order sequences. Prediction accuracy is calculated as a moving average over the last 100 sequences. The sequences are changed after 10,000 elements have been seen (black dashed line). HTM sees each element once, and learns continuously. ELM is trained continuously using a time-lag of 10 steps. LSTM is either retrained every 1000 elements (orange vertical lines) on the last 1000 elements (*yellow*) or 9000 elements (*green*), or continuously adapted using truncated BPTT (*purple*).

A summary of the model performance on the high-order sequence prediction task is shown in Fig. 5. In general, there is a tradeoff between prediction accuracy and flexibility. For batch learning algorithms, a shorter learning window is required for fast adaptation to changes in the data, but a longer learning window is required to perfectly learn high-order sequences (Fig. 5, green vs. yellow). Although online LSTM and ELM do not require batch learning, it does require the user to specify the maximal lag, which limits the maximum sequence order it can learn. The HTM sequence memory model dynamically learns high-order sequences without requiring a learning window or a maximum sequence length. It achieved the best final prediction accuracy with a small number of data samples. After the modification to the sequences, HTM's recovery is much faster than ELM and LSTM trained with batch learning, demonstrating its ability to adapt quickly to changes in data streams.

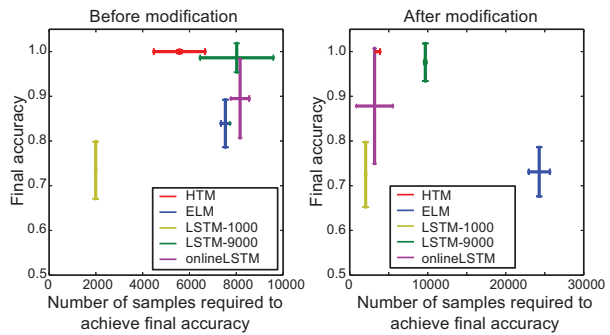

**Figure 5** Final prediction accuracy as a function of the number of samples required to achieve final accuracy before (left) and after (right) modification of the sequences. Error bars represent standard deviations.

### 4.3. Simultaneous multiple predictions

In the experiment with multiple predictions (Fig. 3), each sequence in the dataset has 2 or 4 possible endings, given its high-order context. The HTM sequence memory model rapidly achieves perfect prediction accuracy for both the 2-predictions and the 4-predictions cases. While only these two cases are shown, in reality HTM is able to make many multiple predictions correctly if the dataset requires it. Given a large learning window, LSTM is able to achieve good prediction accuracy for the 2-predictions case, but when the number of predictions is increased to 4 or greater, it is not able to make accurate predictions.

HTM sequence memory is able to simultaneously make multiple predictions due to its use of SDRs. Because there is little overlap between two random SDRs, it is possible to predict a union of many SDRs and classify a particular SDR as being a member of the union with low chance of a false positive (Ahmad and Hawkins, 2016). On the other hand, the real-valued dense distributed encoding used in LSTM is not suitable for multiple predictions, because the average of multiple dense representations in the encoding space is not necessarily close to any of the component encodings, especially when the number of predictions being made is large. The problem can be solved by using local one-hot representations to code target inputs, but such representations have very limited capacity and do not work well when the number of possible inputs is large or unknown upfront. This suggests that modifying LSTMs to use SDRs might enable better performance on this task.

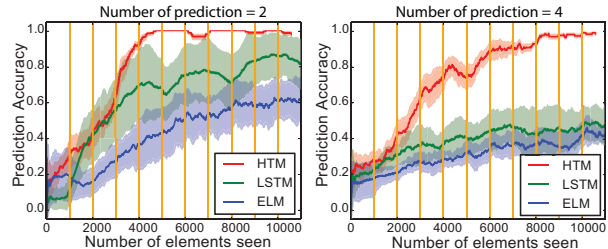

**Figure 6** Performance on high order sequence prediction tasks that require two (left) or four (right) simultaneous predictions. Shaded regions represent standard deviations (calculated with different sets of sequences). The dataset contains four sets of $6^{th}$ order sequences and four sets of $7^{th}$ order sequences.

### 4.4. Learning long term dependencies from high-order sequences

For feedforward networks like ELM, the number of time lags that can be included in the input layer significantly limits the maximum sequence order a network can learn. The conventional recurrent neural networks cannot handle sequences with long term dependencies because error signals "flowing backwards in time" tend to either blow up or vanish with the classical back-propagation through time (BPTT) algorithm. LSTM is capable of learning very long-term dependencies using gating mechanisms (Henaff et al., 2016). Here we tested whether HTM sequence memory can learn long-term dependencies by varying the Markov order of the sequences, which is determined by the length of shared subsequences (Fig. 3A).

We examined the prediction accuracy over training while HTM sequence memory learns variable order sequences. The model is able to achieve perfect prediction performance up to 100-order sequences (Fig. 7A). The number of sequences that are required to achieve perfect prediction performance increases linearly as a function of the order of sequences (Fig. 7B). Note that the model quickly achieves 50% accuracy much faster because it requires only first-order knowledge, yet it requires high-order knowledge to make perfect prediction (Fig. 3A).

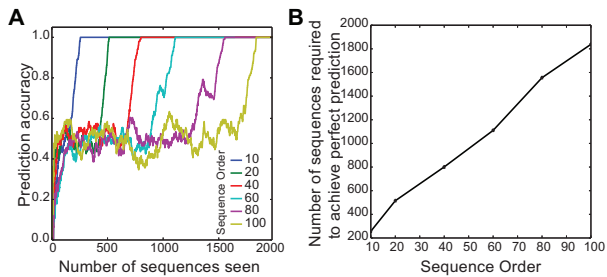

**Figure 7 A**. Prediction accuracy over learning with sequences of different orders. **B**. Number of sequences required to achieve perfect prediction as a function of sequence order. The sequence dataset contains 4 high-order sequences with the structure shown in Fig. 3A.

### 4.5. Disruption of high-order context with temporal noise

In the previous experiments noise was presented between sequences. In this experiment, we tested the effect of noise within sequences. At run time, we randomly replaced either the second, third or fourth element in each sequence with a random symbol. Such temporal noise could disrupt the high-order sequence context and make it much harder to predict the sequence endings. We considered two scenarios: (1) temporal noise throughout training; (2) noise introduced only after the models achieved perfect performance.

The performance of HTM and LSTM are shown in Fig. 8. If temporal noise is present throughout training, neither HTM nor LSTM can make perfect predictions (Fig. 8A). LSTM has slightly better performance than HTM in this scenario, presumably because the gating mechanisms in LSTM can maintain some of the high-order sequence context. HTM behaves like a first order model and has an accuracy of about 0.5. This experiment demonstrates the sensitivity of the HTM model to temporal noise.

If we inject temporal noise after the models achieved perfect performance on the noise-free sequences, the performance of both models drop rapidly (Fig. 8B). The performance of HTM drops to 0.5 (performance of the first-order model), whereas LSTM has worse performance. This result demonstrates that if the high-order sequence context is disrupted, HTM would robustly behave as a low-order model, whereas the performance of LSTM is dependent on the training history.

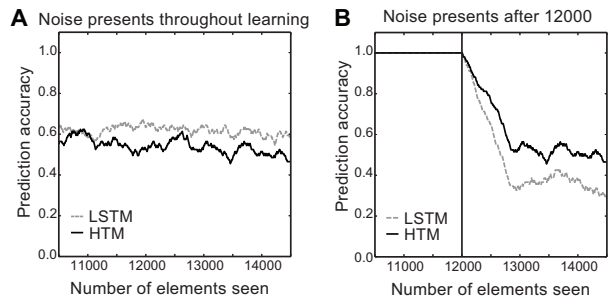

**Figure 8 A.** Prediction accuracy over learning with the presence of temporal noise for LSTM (blue) and HTM (green). **B.** HTM and LSTM are trained with clean sequences. Temporal noise were added after 12000 elements. The sequence dataset is same as in Fig. 4.

### 4.6. Robustness of the network to damage

We tested the robustness of ELM, LSTM and HTM network with respective to removal of neurons. This fault tolerance property is important for hardware implementations of neural network models. After the models achieved stable performance on the high-order sequence prediction task (at the black dashed line, Fig. 2), we eliminated a fraction of the cells and their associated synaptic connections from the network. We then measured the prediction accuracy of both networks on the same data streams for an additional 5000 steps without further learning. There is no impact on HTM sequence memory model performance at up to 30% cell death, whereas performance of ELM and LSTM network declined rapidly with small fraction of cell death (Fig. 9).

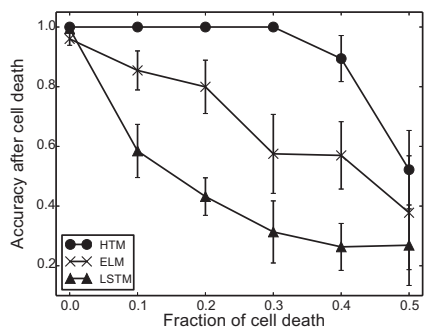

**Figure 9** Robustness of the network to damage. The prediction accuracy after cell death is shown as a function of the fraction of cells that were removed from the network.

Fault tolerance of traditional artificial neural networks depends on many factors, such as the network size and training methods (Lee et al., 2014). The experiments here applied commonly used training methods for ELM and LSTM (see Appendix). It is possible that the fault tolerance of LSTM or any other artificial neural network may be improved by introducing redundancy (replicating trained network) (Tchernev et al., 2005) or by special training method such as dropout (Hinton et al., 2012). In contrast, the fault tolerance of HTM is naturally derived from properties of sparse distributed representations (Ahmad and Hawkins, 2016), in analogy to biological neural networks.

## 5. Prediction of taxi passenger demand

In order to compare the performance of HTM sequence memory with other sequence learning techniques in real-world scenarios, we consider the problem of predicting taxi passenger demand. Specifically, we aggregated the passenger counts in New York City taxi rides at 30-minute intervals using a public data stream provided by the New York City Transportation Authority[2]. This leads to sequences exhibiting rich patterns at different time scales (Fig. 10A). The task is to predict the taxi passenger demand 5 steps (2.5 hours) in advance. This problem is an example of a large class of sequence learning problems that require rapid processing of streaming data to deliver information for real-time decision making (Moreira-Matias et al., 2013).

We applied HTM sequence memory and other sequence prediction algorithms to this problem. The ARIMA model is a widely used statistical approach for time series analysis (Hyndman and Athanasopoulos, 2013). As before, we converted ARIMA and LSTM to an online learning algorithm by re-training the model on every week of data with a buffered dataset of the previous 1000, 3000, or 6000 samples, and by re-training ARIMA at every time step on a buffered dataset of 6000 samples. ELM and ESN were adapted at every time step using sequential online learning methods. The parameters of the ESN, ELM and LSTM network were extensively hand-tuned to provide the best possible accuracy on this dataset. The ARIMA model was optimized using R's "auto ARIMA" package (Hyndman and Khandakar, 2008). The HTM model did not undergo any parameter tuning – it uses the same parameters that were used for the previous artificial sequence task.

We used two error metrics to evaluate model performance: mean absolute percentage error (MAPE), and negative log-likelihood. The MAPE metrics focus on the single best point estimation, while negative log-likelihood evaluates the models' predicted probability distributions of future inputs (see Appendix for details). We found that the HTM sequence memory had comparable performance to LSTM on both error metrics. Both techniques had much lower error than ELM, ESN and ARIMA (Fig. 10B). Note that HTM sequence memory achieves this performance with a single-pass training paradigm, whereas LSTM require multiple-passes on a buffered dataset.

---

[2] http://www.nyc.gov/html/tlc/html/about/trip_record_data.shtml

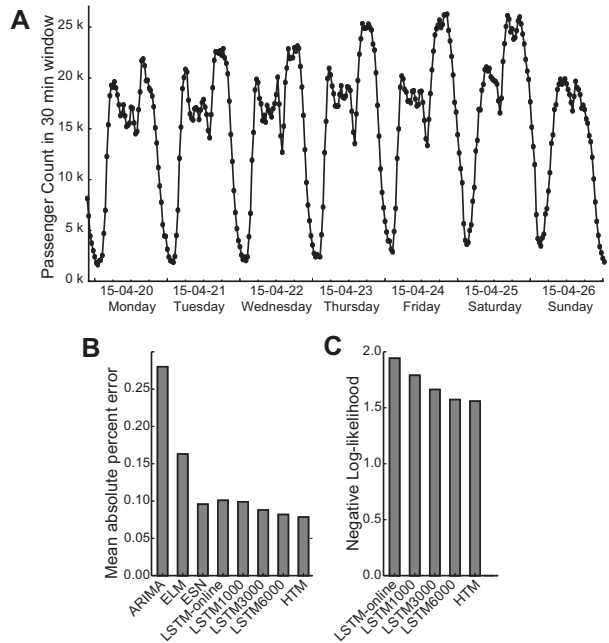

**Figure 10** Prediction of the New York City taxi passenger data. **A**. Example portion of taxi passenger data (aggregated at 30 min intervals). The data has rich temporal patterns at both daily and weekly time scales. **B-C.** Prediction error of different sequence prediction algorithms using two metrics: mean absolute percentage error (B), and negative log-likelihood (C).

We then tested how fast different sequence learning algorithms can adapt to changes in the data. We artificially modified the data by decreasing weekday morning traffic (7am-11am) by 20% and increasing weekday night traffic (9pm-11pm) by 20% starting from April 1$^{st}$. These changes in the data caused an immediate increase in prediction error for both HTM and LSTM (Fig. 11A). The prediction error of HTM sequence memory quickly dropped back in about two weeks, whereas the LSTM prediction error stayed high for a much longer period of time. As a result, HTM sequence memory had better prediction accuracy than LSTM and other models after the data modification (Fig. 11B-C).

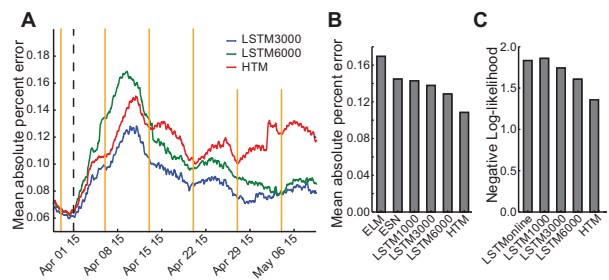

**Figure 11** Prediction accuracy of LSTM and HTM after introduction of new patterns. **A.** The mean absolute percent error of HTM sequence memory (red) and LSTM networks (green, blue) after artificial manipulation of the data (black dashed line). The LSTM networks are re-trained every week at the yellow vertical lines **(B-C)**. Prediction error after the manipulation. HTM sequence memory has better accuracy on both the MAPE and the negative log-likelihood metrics.

## 6. Discussion and conclusions

In this paper we have applied HTM sequence memory, a recently developed neural network model, to real-time sequence learning problems with time-varying input streams. The sequence memory model is derived from computational principles of cortical pyramidal neurons (Hawkins and Ahmad, 2016). We discussed model performance on both artificially generated and real-world datasets. The model satisfies a set of properties that are important for online sequence learning from noisy data streams with continuously changing statistics, a problem the cortex has to solve in natural environments. These properties govern the overall flexibility of an algorithm and its ability to be used in an automated fashion. Although HTM is still at a very early stage compared to other traditional neural network models, it satisfies these properties and shows promising results on real-time sequence learning problems.

### 6.1. Continuous learning with streaming data

Most supervised sequence learning algorithms use a batch-training paradigm, where a cost function, such as prediction error, is minimized on a batch training dataset (Dietterich, 2002; Bishop, 2006). Although we can train these algorithms continuously using a sliding window (Sejnowski and Rosenberg, 1987), this batch-training paradigm is not a good match for time-series prediction on continuous streaming data. A small window may not contain enough training samples for learning complex sequences, while a large window introduces a limit on how fast the algorithm can adapt to changing statistics in the data. In either case a buffer must be maintained and the algorithm must make multiple passes for every retraining step. It may be possible to use a smooth forgetting mechanism instead of hard retraining (Williams and Zipser, 1989; Lughofer and Angelov, 2011), but this requires the user to tune parameters governing the forgetting speed to achieve good performance.

In contrast HTM sequence memory adopts a continuous learning paradigm. The model does not need to store a batch of data as the "training dataset". Instead, it learns from each data point using unsupervised Hebbian-like associative learning mechanisms (Hebb, 1949). As a result the model rapidly adapts to changing statistics in the data.

### 6.2. Using sparse distributed representations for sequence learning

A key difference between HTM sequence memory and previous biologically inspired sequence learning models (Abeles, 1982; Rao and Sejnowski, 2001; Ponulak and Kasiński, 2010; Brea et al., 2013) is the use of sparse distributed representations (SDRs). In the cortex, information is primarily represented by strong activation of a small set of neurons at any time, known as sparse coding (Földiák, 2002; Olshausen and Field, 2004). HTM sequence memory uses SDRs to represent temporal sequences. Based on mathematical properties of SDRs (Kanerva, 1988; Ahmad and Hawkins, 2016), each neuron in the HTM sequence memory model can robustly learn and classify a large number of patterns under noisy conditions (Hawkins and Ahmad, 2016).

A rich distributed neural representation for temporal sequences emerges from computation in HTM sequence memory. Although we focus on sequence prediction in this paper, this representation is valuable for a number of tasks, such as anomaly detection (Lavin and Ahmad, 2015) and sequence classification.

The use of a flexible coding scheme is particularly important for online streaming data analysis, where the number of unique symbols is often not known upfront. It is desirable to be able to change the range of the coding scheme at run-time without affecting the previous learning. This requires the algorithm to use a flexible coding scheme that can represent a large number of unique symbols or a wide range of data. The SDRs used in HTM have a very large coding capacity and allow simultaneous representations of multiple predictions with minimal collisions. These properties make SDR an ideal coding format for the next generation of neural network models.

### 6.3. Robustness and generalization

An intelligent learning algorithm should be able to automatically deal with a large variety of problems without parameter tuning, yet most machine learning algorithms require a task-specific parameter search when applied to a novel problem. Learning in the cortex does not require an external tuning mechanism, and the same cortical region can be used for different functional purposes if the sensory input changes (Sadato et al., 1996; Sharma et al., 2000). Using computational principles derived from the cortex, we show that HTM sequence memory achieves performance comparable to LSTM networks on very different problems using the same set of parameters. These parameters were chosen according to known properties of real cortical neurons (Hawkins and Ahmad, 2016) and basic properties of sparse distributed representations (Ahmad and Hawkins, 2016).

### 6.4. Limitations of HTM and future directions

We have identified a few limitations of HTM. First, as a strict one-pass algorithm with access to only the current input, it may take longer for HTM to learn sequences with very long-term dependencies (Fig. 7) than algorithms that have access to a longer history buffer. Learning of sequences with long-term dependencies can be sped up if we maintain a history buffer and run HTM on it multiple times. Indeed, it has been argued that an intelligent agent should store the entire raw history of sensory inputs and motor actions during interaction with the world (Schmidhuber, 2009). Although it may be computationally challenging to store the entire history, doing so may improve performance given the same amount of sensory experience.

Second, although HTM is robust to spatial noise due to the use of sparse distributed representations, the current HTM sequence memory model is sensitive to temporal noise. It can lose high-order sequence context if elements in the sequence are replaced by a random symbol (Fig. 8). In contrast, the gating mechanisms of LSTM networks appear to be more robust to temporal noise. The noise robustness of HTM can be improved by using a hierarchy of sequence memory models that operate on different time scales. A sequence memory model that operates over longer time scales would be less sensitive to temporal noise. A lower region in the hierarchy may inherit the robustness to temporal noise through feedback connections to a higher region.

Third, the HTM model as discussed does not perform as well as LSTM on grammar learning tasks. We found that on the Reber grammar task (Hochreiter and Schmidhuber, 1997), HTM achieves an accuracy of 98.4% and ELM an accuracy of 86.7% (online training after observing 500 sequences), whereas LSTM achieves an accuracy of 100%. HTM can approximately learn artificial grammars by memorizing example sentences. This strategy could require more training samples to fully learn recursive grammars with arbitrary sequence lengths. In contrast, LSTM learn grammars much faster using the gating mechanisms.

Finally, we have only tested HTM on low-dimensional categorical or scalar data streams in this paper. It remains to be determined whether HTM can handle high-dimensional data such as speech and video streams. The high capacity of the sparse distributed representations in HTM should be able to represent high-dimensional data. However, it is more challenging to learn sequence structure in high-dimensional space, as the raw data could be much less repeatable. It may require additional pre-processing, such as dimensionality reduction and feature extractions, before HTM can learn meaningful sequences with high-dimensional data. It would be an interesting future direction to explore how to combine HTM with other machine learning methods, such as deep networks, to solve high-dimensional sequence learning problems.

# 7. Appendix

## 7.1. HTM sequence model implementation details

In our software implementation, we made a few simplifying assumptions to speed up simulation for large networks. We did not explicitly initialize a complete set of synapses across every segment and every cell. Instead, we greedily created segments on the least used cells in an unpredicted column and initialized potential synapses on that segment by sampling from previously active cells. This happened only when there is no match to any existing segment. The initial synaptic permanence for newly created synapses is set as 0.21(Table 1), which is below the connection threshold (0.5).

The HTM sequence model operates with sparse distributed representations (SDRs). Specialized encoders are required to encode real-world data into SDRs. For the artificial datasets with categorical elements, we simply encoded each symbol in the sequence as a random SDR, with 40 randomly chosen active bits in a vector of 2048 bits.

For the NYC taxi dataset, three pieces of information were fed into the HTM model: raw passenger count, the time of day, and the day of week (LSTM received the same information as input). We used NuPIC's standard scalar encoder to convert each piece of information into an SDR. The encoder converts a scalar value into a large binary vector with a small number of ON bits clustered within a sliding window, where the center position of the window corresponds to the data value. We subsequently combined three SDRs via a competitive sparse spatial pooling process, which also resulted in 40 active bits in a vector of 2048 bits as in the artificial dataset. The spatial pooling process is described in detail here (Hawkins et al., 2011).

The HTM sequence memory model used an identical set of model parameters for all the experiments described in the paper. A complete list of model parameters is shown below. The full source code for the implementation is available on Github at https://github.com/numenta/nupic

**Table 1 Model parameters for HTM**

| Parameter name | Value |
| --- | --- |
| Number of columns $N$ | 2048 |
| Number of cells per column $M$ | 32 |
| Dendritic segment activation threshold $\theta$ | 15 |
| Initial synaptic permanence | 0.21 |
| Connection threshold for synaptic permanence | 0.5 |
| Synaptic permanence increment $p^+$ | 0.1 |
| Synaptic permanence decrement $p^-$ | 0.1 |
| Synaptic permanence decrement for predicted inactive segments $p^-$ | 0.01 |
| Maximum number of segments per cell | 128 |
| Maximum number of synapses per segments | 128 |
| Maximum number new synapses added at each step | 32 |

## 7.2. Implementation details of other sequence learning algorithms

*ELM* We used the online sequential learning algorithm for ELM (Liang et al., 2006). The network used 50 hidden neurons and a time lag of 100 for the taxi data and 200 hidden neurons and a time lag of 10 for the artificial dataset.

*ESN* We used the Matlab toolbox for Echo State Network developed by Jaeger's group (http://reservoir-computing.org/node/129). The ESN network 100 internal units, a spectral radius of 0.1, a teacher scaling of 0.01 and a learning rate of 0.1 for the ESN model. The parameters were hand tuned to achieve the best performance. We used the online learning mode and adapted the weight at every time step.

*LSTM* We used the PyBrain implementation of LSTM (Schaul et al., 2010). For the artificial sequence learning task, the network contains 25 input units, 20 internal LSTM neurons and 25 output units. For the NYC taxi task, the network contains 3 input units, 20 LSTM cells, 1 output units for calculation of the MAPE metric and 22 output units for calculation of the sequence likelihood metric. The LSTM cells have forget gates but not peephole connections. The output units have a biased term. The maximum time lag is the same as the buffer size for the batch-learning LSTMs. We used two training paradigms. For the batch-learning paradigm, the network were retrained every 1000 iterations with a popular version of the resilient backpropagation method (Igel and Hüsken, 2003). For the online-learning paradigm, we calculated the gradient at every time step using truncated backpropgation through time over the last 100 elements (Williams and Peng, 1990), and adjusted the parameters along the gradient with a learning rate of 0.01.

### 7.3. Evaluation of model performance in the continuous sequence learning task

Two error metrics were used to evaluate the prediction accuracy of the model. First, we considered mean absolute percentage error (MAPE) metric, an error metric that is less sensitive to outliers than root mean squared error.

$$\text{MAPE} = \frac{\sum_{t=1}^{N} |y_t - \hat{y}_t|}{\sum_{t=1}^{N} |y_t|} \qquad (1)$$

In Eq. 1, $y_t$ is the observed data at time *t*, $\hat{y}_t$ is the model prediction for the data observed at time *t*, and *N* is the length of the dataset.

A good prediction algorithm should output a probability distribution of future elements of the sequence. However, MAPE only consider the single best prediction from the model, and thus do not incorporate other possible predictions from the model. We used negative log-likelihood as a complementary error metric to address this problem. The sequence probability can be decomposed into:

$$p(y_1, y_2, \ldots, y_t) = p(y_1)p(y_2|y_1)p(y_3|y_1, y_2)p(y_t|y_1, \ldots, y_{t-1}) \qquad (2)$$

The conditional probability distribution is modeled by HTM or LSTM based on network state at the previous time step.

$$p(y_t|y_1, \ldots, y_{t-1}) = P(y_t|\text{network state}_{t-1}) \qquad (3)$$

The negative log-likelihood of the sequence is then given by:

$$NLL = \frac{1}{N} \sum_{t=1}^{N} \log P(y_t|\text{model}) \qquad (4)$$